\documentclass{INTERSPEECH2023}
\usepackage[latin9]{inputenc}
\usepackage{float}
\usepackage{multirow}
\usepackage{amsmath,bm,amssymb}
\usepackage{graphicx}
\usepackage{hyperref}
\usepackage{ragged2e}
\usepackage{microtype}
\usepackage{subfigure}
\usepackage{booktabs} 
\usepackage{tikz}
\usetikzlibrary{positioning}
\justifying

\makeatletter

\providecommand{\tabularnewline}{\\}

\interspeechcameraready

\usepackage{epstopdf}
\usepackage{epsfig}
\usepackage{multicol}
\usepackage{comment}
\usepackage{url}

\title{O-1: Self-training with Oracle and 1-best Hypothesis}
\name{\parbox{0.9\linewidth}{\center Murali~Karthick~Baskar, Andrew~Rosenberg, Bhuvana~Ramabhadran, Kartik Audhkhasi}}
\address{
Google Inc., 
}
\email{\{mkbaskar,rosenberg,bhuv\}@google.com}

\makeatother

\begin{document}
\ninept \maketitle 
\begin{abstract}
We introduce O-1, a new self-training objective to reduce training bias and unify training and evaluation metrics for speech recognition. O-1 is a faster variant of Expected Minimum Bayes Risk (EMBR), that boosts the oracle hypothesis and can accommodate both supervised and unsupervised data. We demonstrate the effectiveness of our approach in terms of recognition on publicly available SpeechStew datasets and a large-scale, in-house data set. On Speechstew, the O-1 objective closes the gap between the actual and oracle performance by 80\% relative compared to EMBR which bridges the gap by 43\% relative. O-1 achieves 13\% to 25\% relative improvement over EMBR on the various datasets that SpeechStew comprises of, and a 12\% relative gap reduction with respect to the oracle WER over EMBR training on the in-house dataset. Overall, O-1 results in a 9\% relative improvement in WER over EMBR, thereby speaking to the scalability of the proposed objective for large-scale datasets. 

\end{abstract}


\section{Introduction}


The use of labeled and unlabeled data in training has led to a suite of co-training and self-training~\cite{wei2021theoretical} methods in speech and language processing. The underlying principle in these algorithms is an iterative learning paradigm where a model learns from a ground-truth distribution as well as from its own predictions at the previous iteration with appropriate regularization~\cite{grandvalet2004semi}. These algorithms serve to mitigate three key problems. First, \textit{exposure bias} in autoregressive models~\cite{RanzatoCAZ15}, where the model is conditioned only from the supervised training data (teacher forcing)~\cite{williams1989learning} 
and thus never learns from its own predictions. This leads to early convergence but is mismatched with the inference step, where no ground-truth is available. Scheduled sampling~\cite{bengio2015scheduled} mitigates this issue by exposing the model to its past/own predictions at random instances.  
Second, \textit{mismatches between training objectives and evaluation metrics} such as, maximum likelihood estimation (MLE) vs Word Error Rate (WER))~\cite{Bahdanau2016AnAA,RanzatoCAZ15} which cause performance degradation of the model.  
While the former results in error accumulation during inference, the latter ignores the lattice of predictions which can be utilized to learn from unsupervised data.
Thirdly, \textit{label bias} in e2e systems occurs when the model cannot relearn from new labels and remains biased to past labels. This label bias is mostly subtle and is studied and addressed in~\cite{pmlr-v97-collobert19a} with the use of global normalization.

Prior work on Automated Speech Recognition (ASR) has shown that, Minimum Bayes risk (MBR) approach is capable of handling the exposure bias~\cite{prabhavalkar2018minimum,lu21b_interspeech}, training-inference criterion mismatch~\cite{povey2005discriminative,muller-sennrich-2021-understanding} and the label bias problem~\cite{pmlr-v97-collobert19a,lu2021input}.
MBR is a self-training paradigm (commonly referred to as discriminative training)~\cite{goel2000minimum} and is either termed as expected MBR (EMBR)~\cite{shannon2017optimizing,mkbaskarprefix,weiran22_interspeech} when used to reduce expected word errors or state MBR (sMBR) when minimizing the state level expected errors~\cite{vesely2013sequence,povey2001improved, povey2005discriminative,kingsbury2009lattice,povey2007evaluation}.
Despite the benefits of EMBR, it suffers from two main drawbacks:
\begin{itemize}
    \item EMBR formulation does not directly impact the oracle hypothesis (best WER), but instead focusses on expected error reduction by boosting multiple hypotheses scores.
    \item The training duration is relatively high compared to MLE training as  batch size gets scaled by the beam size.
\end{itemize}
In this work, we propose a novel self-training objective, referred to as O-1. It is a variant of EMBR formulated to discriminatively boost only the \textbf{O}racle hypothesis from the \textbf{1}-best hypothesis. 
The primary contributions of this work are:
\begin{itemize}
    \item We introduce, the O-1 objective which allows the efficient use of both supervised data and unsupervised data. O-1 is integrated with hard distillation~\cite{fsdistill} for unsupervised training. To the best of our knowledge, O-1 is the first variant of EMBR which can perform unsupervised learning.  
    \item Optimized training time (and computational costs) compared to EMBR, thereby paving the way to train for longer duration and more data. 
    \item Effective self-training paradigm that uses WER to to select the best sequence predicted by the model as the target label sequence.
    \item The proposed objective results in a significant reduction of the gap between 1-best WER and the  oracle WER. Experiments conducted on the set of public corpora in Speechstew show consistent reduction in WER, by 13-25 \% relative.
\end{itemize}

\section{Related Work}
Exposure bias~\cite{bengio2015scheduled, rennie2017self,he2019exposure} is well-known as a central problem for auto-regressive models such as language models. Scheduled sampling~\cite{bengio2015scheduled} is a form of self-training proposed to alleviate this training-inference bias~\cite{arun2010unified} caused by the maximum-likelihood (MLE) based training objective. Beam search optimization was proposed to reduce exposure bias in ~\cite{wiseman-rush-2016-sequence}. It collects and improves the scores of each ground truth token and the previous token leading to incorrect prediction in each beam search step. Recent work on reducing exposure bias for ASR using recurrent neural network transducers (RNN-T) claims that the direct application of scheduled sampling to RNN-T architectures is challenging as it is not a generative architecture~\cite{cui21b_interspeech}, unlike self-attention based transformer like encoder-decoder architectures~\cite{bahdanau2016end}. To mitigate this bias, the authors sample the target label from the top-k candidates from a language model instead of the model itself. 
In~\cite{Guo2020}, the EMBR computation is optimized for RNN-T architectures by using alignment scores for target hypotheses. The bias introduced during training-inference with LM rescoring is addressed in~\cite{weiran22_interspeech} using the EMBR objective. Prefix level EMBR is proposed in~\cite{mbrdec2003,sabour2018optimal,mkbaskarprefix} to impose correct prefixes in the beam by boosting prefixes with less errors. Scheduled sampling uses a single hypothesis and  EMBR uses multiple hypotheses in their objectives, but neither of the above approaches select the oracle hypothesis and boosts its score. The proposed O-1 mitigates this constraint and has shown  consistent improvements in recognition performance.
\begin{figure}[!tbh]
    \centering
    \includegraphics[scale=0.24]{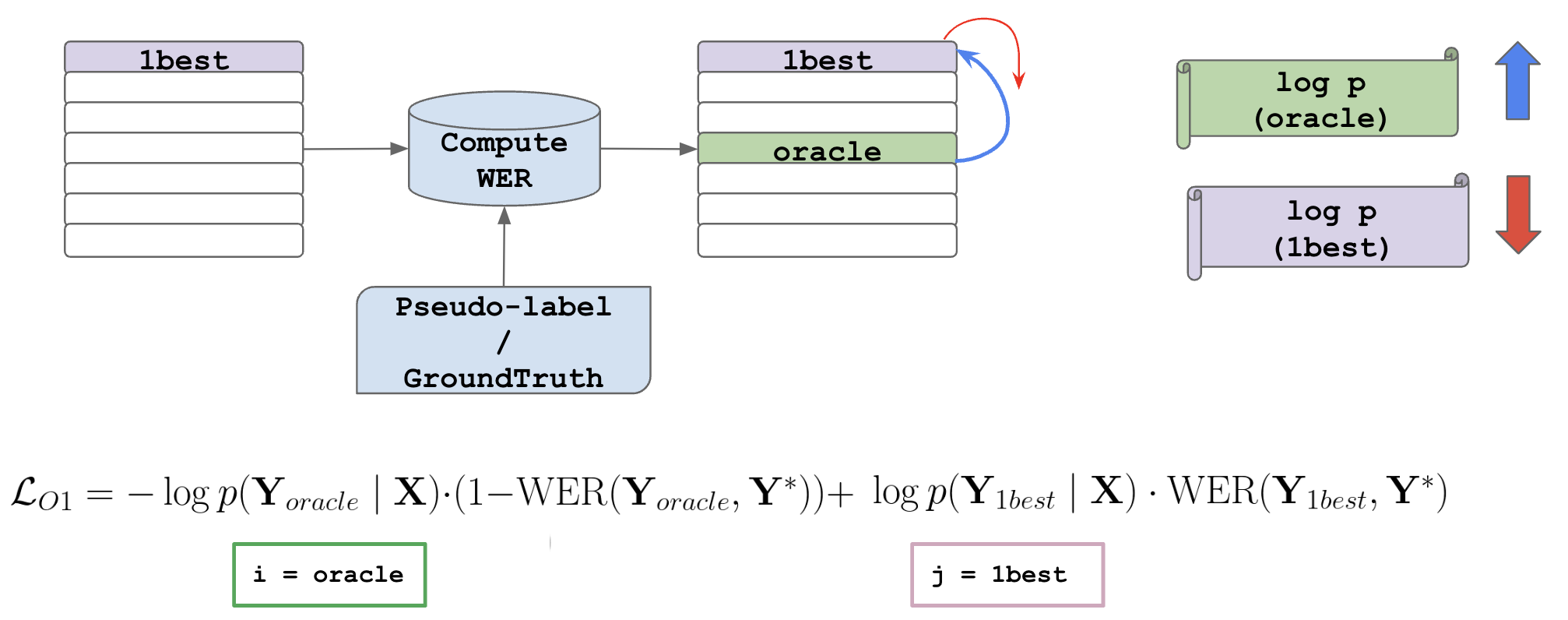}
    \caption{Visualization of O-1 training defined in Equation~\eqref{eq:o1} that boosts the score of the oracle hypothesis to the top of the beam.}
    \label{fig:o1}
\end{figure}
\vspace{-0.7cm}
\section{O-1: Proposed Approach}\label{sec:o1}
In this work, we use Recurrent Neural Network Transducers~\cite{graves2012sequence} are used as the  framework for EMBR and the proposed O-1 objectives. Figure~\ref{fig:o1} visualizes the training procedure corresponding to Equation~\eqref{eq:o1}. 
\subsection{EMBR}
Given an input speech sequence $\mathbf{X}$, the encoder network, prediction and joint network in an RNN-T model help to compute the alignment probability $p(\mathbf{a} \mid \mathbf{X})$ with the help of forward-backward algorithm. The RNN-T training objective $\mathcal{L}_{rnnt}$ is defined as:
\begin{equation}
\begin{split}
    \mathcal{L}_{rnnt} = - \log p(\mathbf{Y} \mid \mathbf{X}) = - \log\sum_{\mathbf{a} \in \mathcal{B}^{-1}(\mathbf{Y})}p(\mathbf{a} \mid \mathbf{X})
\end{split}\label{eq:rnnt}
\end{equation}
Here, $\mathcal{B}$ operation removes all blank labels in a given sequence and $p(\mathbf{Y}\mid\mathbf{X})$ is the probability of the predicted hypothesis $\mathbf{Y}$.
Based on the objective in Equation~\eqref{eq:rnnt}, the EMBR loss~\cite{Guo2020} for RNNT is computed as:
\begin{equation}
    \begin{split}
        \mathcal{L}_{embr} = \sum_{\mathbf{Y}_i \in \mathcal{Y}} \hat{p} (\mathbf{Y}_{i} \mid \mathbf{X}) W(\mathbf{Y}_i, \mathbf{Y}^{*})\label{eq:embr}
    \end{split}
\end{equation}
where $\hat{p}(\mathbf{Y}_{i} \mid \mathbf{X})=\textrm{softmax}(p(\mathbf{Y}_{i}\mid\mathbf{X}))$ is the probability of $i^{\textrm{th}}$ hypothesis in the beam obtained by applying a softmax function on $\log p$ of the RNN-T model to normalize over the n-best hypotheses. Here, $\mathcal{Y}$ denotes the n-best list and $W(\mathbf{Y}_i, \mathbf{Y}^{*})$ is the number of erroneous words computed using an edit distance between each hypothesis in the beam $\mathbf{Y}_{i}$ and the ground-truth $\mathbf{Y}^{*}$. 
In the partial derivative of the EMBR objective below,
\begin{equation}
\begin{split}
    \frac{\partial \mathcal{L}_{embr}}{\partial \log p(\mathbf{Y}_{i} \mid \mathbf{X})} 
    &= \hat{p}(\mathbf{Y}_{i}\mid \mathbf{X}) (W(\mathbf{Y}_{i}, \mathbf{Y}^{*}) - \bar{W}) \\
\end{split}\label{eq:gradembr}
\end{equation}
$\bar{W}=\underset{\mathbf{Y}_{j} \in \textrm{n-best}}{\sum} \hat{p} (\mathbf{Y}_{j} \mid \mathbf{X}) W(\mathbf{Y}_{j}, \mathbf{Y}^{*})$ denotes the expected word errors or risk among all the hypotheses for training an utterance. Monte-Carlo approximation of the n-best list results in $\bar{W}\approx\underset{\mathbf{Y}_{j} \in \textrm{n-best}}{\sum}W(\mathbf{Y}_{j}, \mathbf{Y}^{*})$ as the average number of word errors. The subtraction of $\bar{W}$ in ~\eqref{eq:gradembr} performs variance reduction~\cite{sutton1999policy,shannon2017optimizing}. In~\cite{prabhavalkar2018minimum}, the authors  applied $\bar{W}$ directly in ~\eqref{eq:embr} also as a form of variance reduction, but computing $\bar{W}$ is empirical and tricky~\cite{rennie2017self}.   

\subsection{O-1 Training Procedure}
The O-1 training objective $\mathcal{L}(\mathbf{X}, \mathbf{Y}, \mathbf{Y}^{*})$ is defined as:
\begin{equation}
\begin{split}
    \mathcal{L}_{O1} = - \log p (\mathbf{Y}_{oracle} \mid \mathbf{X}) \cdot (1 - \textrm{WER}(\mathbf{Y}_{oracle}, \mathbf{Y}^{*})) \\
    + \log p (\mathbf{Y}_{1-best} \mid \mathbf{X}) \cdot \textrm{WER}(\mathbf{Y}_{1-best}, \mathbf{Y}^{*})\label{eq:o1}
\end{split}
\end{equation}
The training steps to implement Equation~\eqref{eq:o1} are enumerated below:
\begin{enumerate}
    \item Beam-search decoding is performed during training to extract  n-best hypotheses $\mathbf{Y}_{i=1:n}$.
    \item Each hypothesis $\mathbf{Y}_i$ is scored against the ground-truth or pseudo-label from the teacher to compute WER. 
    \item The hypothesis with the best WER serves as the oracle hypothesis $\mathbf{Y}_{oracle}$ and the hypothesis with the top probability becomes the 1-best hypothesis $\mathbf{Y}_{1-best}$. Both oracle and 1-best hypotheses are chosen from the beam while dropping the rest of the hypotheses. 
    \item $- \log p(\mathbf{Y}_{i,j} \mid \mathbf{X})$ is the RNN-T loss for $i=oracle$ hypothesis and $j=1-best$ hypothesis computed using ~\eqref{eq:rnnt}.
    \item Each of these two RNN-T losses is scaled by its corresponding WER as described in Equation~\eqref{eq:o1}. This idea of scaling objectives with the evaluation metric has also been explored for pre-training in ~\cite{baskar2022ask2mask}.
    \item When the scores of 1-best and oracle hypotheses are equal, $\mathcal{L}_{O1} \rightarrow 0$  in ~\eqref{eq:o1} and the auxiliary $\mathcal{L}_{rnn-t}$ loss will overcome this zero loss scenario as given in Equation ~\eqref{eq:embraux}.
\end{enumerate}
It can be seen that O-1 training simplifies EMBR training by eliminating the need to consider all the alternate hypotheses and eliminates the computation of $\bar{W}$. O-1 objective also does not require a softmax computation that EMBR requires (Equation~\eqref{eq:embr}).
The final EMBR and O-1 losses used in our experiments is a multitask training objective given by:
\begin{equation}
    \begin{split}
        \mathcal{L}_{embr}^{'} = \mathcal{L}_{embr} + \gamma \mathcal{L}_{rnnt} \\
        \mathcal{L}_{O1}^{'} = \mathcal{L}_{O1} + \lambda \mathcal{L}_{rnnt}\label{eq:embraux}
    \end{split}
\end{equation}
O-1 explicitly targets the hypothesis with less WER to be boosted without relying on a error threshold $W - \bar{W}$ as in Equation~\eqref{eq:gradembr}. O-1 performs discriminative training between oracle and 1-best scores. This formulation of O-1 naturally addresses the label bias problem as well; when the model is pushing the oracle hypothesis score to be on top of the beam, the 1-best hypothesis score has to come down to respect the discriminative aspect of O-1. 

\subsection{O-1 Unsupervised training}
Hard distillation~\cite{park2020improved,hwang2022comparison} involves training a student model with the target labels obtained on-the-fly from the teacher model as in~\eqref{eq:distill}. 
Given the reduced computational complexity compared to EMBR, we choose to employ O-1 in an unsupervised setting using the distillation loss given in ~\cite{park2020improved}.
Thus,  O-1 loss with hard distillation becomes:
\begin{equation}
\begin{split}
    \mathcal{L}_{distill} &= \mathcal{L}_{rnnt}(\mathbf{X}, \mathbf{Y}^{'}) \\
    \mathcal{L}_{O1\_distill} &= \mathcal{L}_{O1}(\mathbf{X}, \mathbf{Y}, \mathbf{Y}^{'}) + \lambda \, \mathcal{L}_{distill},\label{eq:distill}
\end{split}    
\end{equation}
where, $\mathbf{Y}^{'}$ denotes the pseudo-label generated by the teacher model during training.

\section{Experiments}
\subsection{Datasets}
We conduct experiments on both publicly available and in-house corpora. SpeechStew, described in~\cite{chan2021speechstew}, a publicly available and well-benchmarked corpus, composed of multiple datasets ($\approx$5000 hours) spanning several acoustic conditions and domains. It comprises of 960 hrs of Librispeech, 100 hrs of AMI, 1.5k hrs of CommonVoice corpus, 50 hrs of English broadcast news, 2k hrs of Switchboard/Fisher, 450 hrs of TED-lium and 80 hrs of WSJ corpus.
The in-house corpus consists of 
Google's 
short 
Voice Search (VS)
 utterances in U.S. English. This \textit{supervised} ASR training data  is anonymized and hand-transcribed. The development and test sets are a small fraction of the training set held out for validation and evaluation. Also included in the training are nearly 300 thousand hours of multi-domain utterances spanning domains
 such as  Dictation, YouTube and Telephony
 for US English~\cite{narayanan2019recognizing}. The \textit{unsupervised} dataset contains 500M short 
 search 
 utterances. While the public corpora allows us to benchmark against state-of-the-art algorithms, the in-house corpora allows us to study performance on tail distributions and evaluate the scalability of the algorithm to large-scale training data and streaming E2E setups.  

\subsection{Model Description}
The model architecture used in this work is based on the RNN-T architecture first described in~\cite{graves2012sequence}. It consists of a conformer encoder~\cite{gulati2020conformer} and an LSTM decoder. 
The encoder comprises of several blocks, with each block including a series of multi-headed
self attention, depth-wise convolution and feed-forward layers. For our experiments with SpeechStew, we use two different configurations, namely a 100M and 600M parameter model with 17 and 24 conformer layers with 512 and 1024 encoder dimensions, respectively. Both models use relative attention.  All models are trained on use 80-dimensional log-mel filter bank coefficients. We use SpecAugment~\cite{park2019specaugment} with time and frequency masking for all experiments. For SpeechStew, the number of frequency masking bins $F$ = 27 and time bins $T$ = 10 with a maximum mask ratio set to 0.05. For the in-house dataset, $F$ = 27, $T$ = 50 and mask ratio to 0.1.
For experiments with the in-house corpus, we use a 120M parameter, streaming Hybrid Autoregressive Transducer (HAT)~\cite{variani2020hybrid} architecture, comprising of a conformer encoder with 12 layers 
and a step decoder with V2 embedding as described  in~\cite{weiran22_interspeech}. 
In addition to SpecAugment, we use multistyle training similar to ~\cite{kim2017generation}.  The experiments on SpeechStew use 1024 word-piece targets and the in-house experiments on the Voice Search task use 4K word-piece targets derived using the algorithm presented. in~\cite{kudo2018sentencepiece}.


\subsection{O-1 Supervised Self Training}

We use the following training strategy to study the impact of the proposed algorithm with supervised-only training data. A 100M RNN-T model is trained from scratch using the supervised, SpeechStew corpora. The 600M parameter model is finetuned from a wav2vec2~\cite{BaevskiZMA20} encoder that is pretrained with Librilight data. The model is trained to convergence (approximately 300K iterations) and serves as the baseline model for subsequent EMBR and O-1 training (100K iterations). Based on empirical evidence we set the RNN-T loss scaling factor as $\lambda=0.1, \gamma=0.1$ for the EMBR $\mathcal{L}_{embr}$ and the O-1 training objectives $\mathcal{L}_{O1}$. In all our O-1 experiments, the RNN-T loss in Equation~\eqref{eq:rnnt} is normalized with the target label sequence length before using for O-1 loss computation in~\eqref{eq:o1}. SpeechStew test sets~\cite{chan2021speechstew} are used for evaluation.

\subsection{O-1 Unsupervised Learning}
In the unsupervised learning scenario, the teacher is a non-causal model with right context fixed to 64 and the student is a streaming model with zero right context. The teacher model is trained with the supervised data alone and the baseline student model is trained with the same supervised data and additional unsupervised data. 
For unsupervised learning, we use a teacher model's transcripts as pseudo labels. We use the hard distillation (Equation ~\eqref{eq:distill}), to train both student and teacher models jointly. Typically CE training converges after 500K iterations. This serves as the CE baseline to continue with EMBR training (Equation ~\eqref{eq:embr}). For O-1 training (Equation ~\eqref{eq:o1}), the teacher and student are initialized in the same way the baseline, i.e., we do not require the additional CE training step that conventional EMBR training requires. As explained in Section~\ref{sec:o1}, CE objective is included as an auxiliary loss in the O-1 objective. All experiments related to unsupervised learning are conducted on the in-house corpora,  since vast amounts of untranscribed data is available. Each training batch contains equal amounts of transcribed and untranscribed data. Beam-search is done with a beam size of 8 for hypotheses generation in EMBR training and inference. In the case of O-1, beam size of 18 is set for SpeechStew and 8 for in-house dataset during training. O-1 inference is done using beam size of 8 to perform fair comparison with EMBR.
\begin{figure*}[ht]
    \centering
    \includegraphics[trim={0cm 0cm 0cm 1cm},clip,scale=0.30]{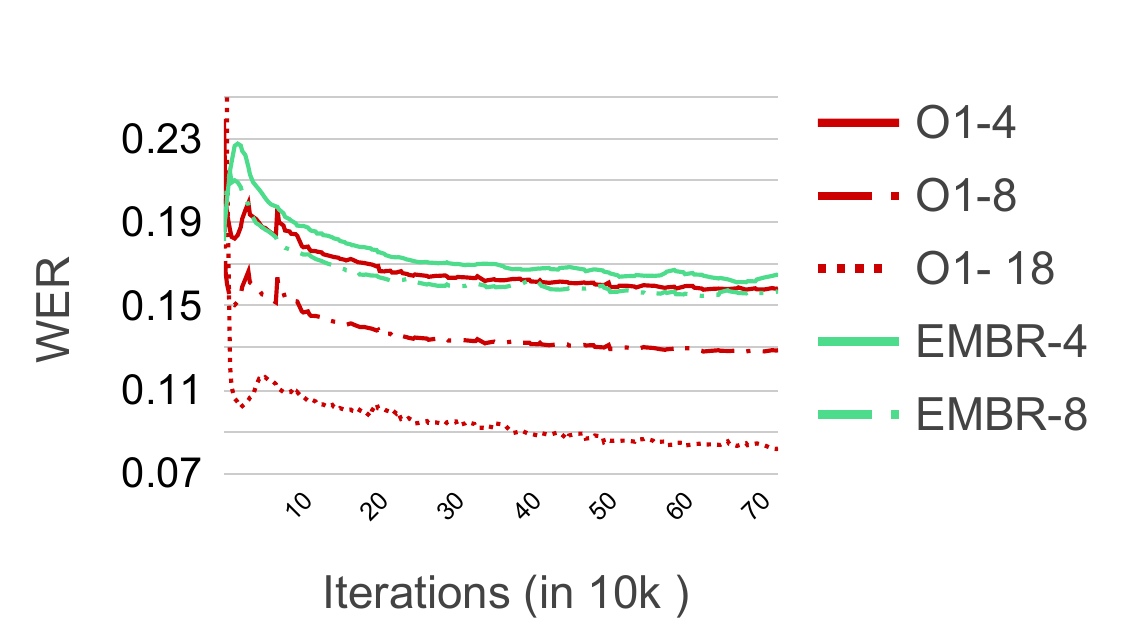}
    \includegraphics[scale=0.30]{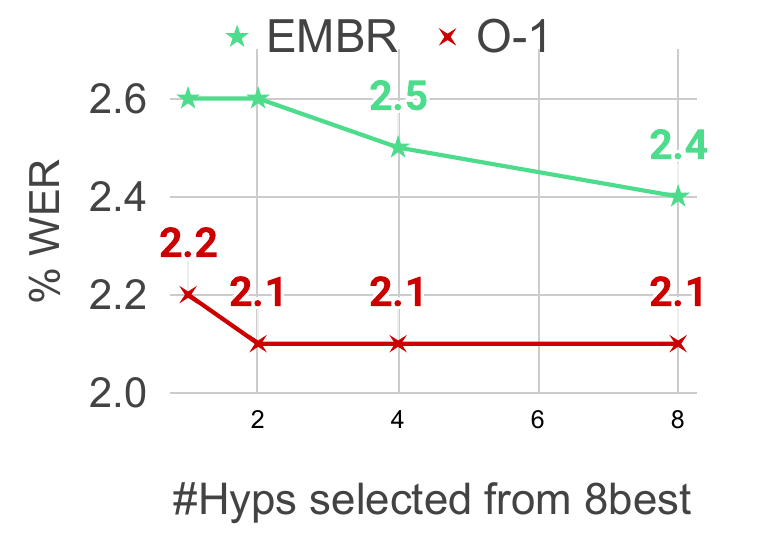}
    \includegraphics[trim={0cm 0cm 1cm 1cm},scale=0.29]{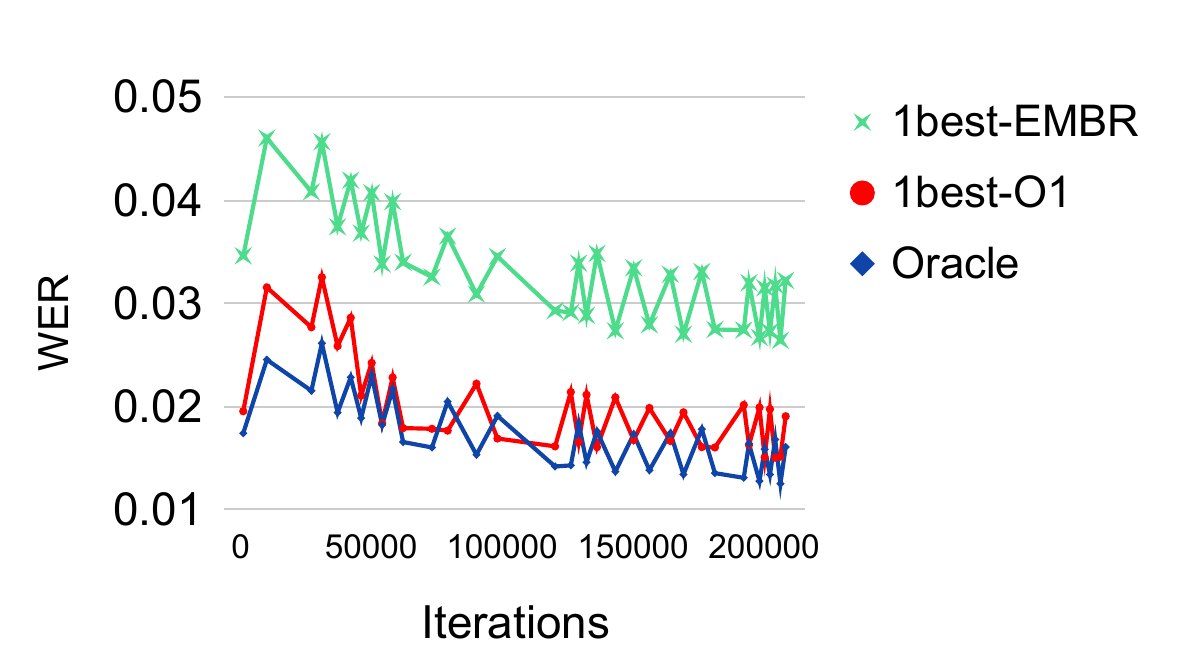}
   \caption{[A] Average WER of all test sets in SpeechStew evaluated at different checkpoints trained with different beam sizes using the proposed O-1 objective and EMBR. [B] Plot showing the number of hypotheses required to reach the optimal performance in EMBR and O-1. Beam search is fixed to compute 8 hypotheses. EMBR is computed from top8 hypotheses. O-1's  WER is close to the oracle WER (1.67\%) compared to EMBR on librispeech test clean. Model is trained on SpeechStew datasets. [C] 1-best WER vs oracle WER on Librispeech test-clean  using O-1, baseline and EMBR}
    \label{fig:numhyp}
\end{figure*}
\section{Results and Analysis}
Table~\ref{tab:stew} shows the performance comparison of O-1 with the baseline and the EMBR models. EMBR shows inconsistent behavior in different testsets. For instance, EMBR degrades on AMI-IHM, Librispeech \textit{test} and \textit{testother}, TED-LIUM and WSJ-eval92. EMBR shows gains on AMI-SDM, CHiME and Commonvoice which are relatively harder tasks over rest of the datasets. O-1 provides consistent improvement in all testsets over both the baseline and the EMBR models with a maximum of 25\% relative gain on WSJ-eval92. The primary reason for O-1's  improved performance is learning from the best hypothesis targets. This enables the model to achieve closer to oracle performance during training as shown in Figure~\ref{fig:numhyp}C and gets reflected in evaluation.
\begin{table}[ht]
\centering
\caption{Recognition performance on Speechstew evaluation sets with 100M and 600M parameter models}
\label{tab:stew}
\resizebox{8.3cm}{1.2cm}{
\begin{tabular}{ c c c c c c c c c c }
\toprule 
\multirow{2}{*}{\#Params} & \multirow{2}{*}{\% WER} & \multicolumn{2}{c}{AMI} & \multirow{2}{*}{CHiME} & \multirow{2}{*}{CV} & \multicolumn{2}{c}{Libri-test} & \multirow{2}{*}{TED} & \multirow{2}{*}{WSJ}\tabularnewline
\cline{3-4} \cline{4-4} \cline{7-8} \cline{8-8} 
 & & IHM & SDM &  &  & clean & other &  & \tabularnewline
\hline 
\midrule 
\multirow{3}{*}{100M} & Baseline & 10.2 & 24.8 & 34.9 & 11.5 & 2.3 & 4.9 & 4.8 & 2.6\tabularnewline
 & EMBR & 11.3 & 23.7 & 33.2 & 11.2 & 2.4 & 5.4 & 4.7 & 2.7\tabularnewline
 & O-1 & 9.8 & 23.6 & 31.6 & 11.1 & 2.1 & 4.6 & 4.4 & 2.0\tabularnewline
\midrule
\multirow{3}{*}{600M} & Baseline & 9.6 & 23.8 & 33.5 & 11.5 & 1.8 & 3.6 & 5.6 & 1.4 \tabularnewline
 & EMBR & 9.1 & 22.7 & 29.7 & 10.2 & 1.9 & 3.6 & 4.8 & 1.4 \tabularnewline
 & O-1 & 8.7 & 20.4 & 25.3 & 8.9 & 1.7 & 3.4 & 4.3 & 1.3 \tabularnewline
\bottomrule 
\end{tabular}}
\end{table}
We also conducted the same experiment on a 600M parameter model to investigate if the performance of the new objective is complementary to the increase in model complexity. 
Table~\ref{tab:stew} shows that on average, O-1 achieves 8.4\% rel. improvement over EMBR and 13\% relative improvement over the state-of-the-art baseline system in ~\cite{chan2021speechstew}. O-1 shows relative improvements of up to 25\% on WSJ, 15\% on CHiME and greater than 10\% on TED, AMI-SDM and CommonVoice.
\subsection{Comparison with Scheduled Sampling}
Analogous to our approach, scheduled sampling also allows the model to train from its own predictions. However, O-1 differs from scheduled sampling in two aspects:
    (1) The oracle hypothesis serves as the target score to be improved, while the 1-best hypothesis score is suppressed. In the case of scheduled sampling, the 1-best hypothesis is chosen as the target based on the selection ratio.
    (2) The oracle hypothesis score is scaled by the corresponding WER in O-1 which is not done in scheduled sampling.
Empirical analysis is conducted on Librispeech test set to evaluate the behavior of O-1 and scheduled sampling approach. Here we perform scheduled sampling at the utterance level without any language model~\cite{cui21b_interspeech}. O-1 obtains 8\% relative improvement over scheduled sampling approach on test-other and 12.5\% relative gain on test.
\subsection{Training duration, Performance and Beam size}
During O-1 training all the samples in the beam are not involved in loss computation which reduces the exposure to erroneous samples. Batch size is not impacted by the number of samples in the beam during O-1 training. These two factors allow the model tained with O-1 to use a larger beam size unlike EMBR training. Figure~\ref{fig:numhyp}A shows the effect of increasing the beam size during training. "O1 - 18" (O-1 trained with beam size of 18) shows substantial gains over EMBR.
We also conducted an experiment by training using the oracle hypothesis only and this achieved a train time similar to the baseline model. Efficient inferencing on TPUs~\cite{jouppi2021ten} results in maximum computation resting on the N-best hypothesis generation. 
On average, the O-1 processes 945 examples/sec and EMBR processes 754 examples/sec for the SpeechStew data with a total of 3.6 million examples, which makes it faster over EMBR training. 
\subsection{Are all hypothesis needed during training ?}
Figure~\ref{fig:numhyp}B shows the performance of EMBR and O-1 based on the number of hypotheses involved in loss computation. Here, we argue that the reason for EMBR attaining the best performance with 8 hypotheses is that on average the oracle hypothesis is between 5$^\textrm{th}$ and 8$^\textrm{th}$ position within the top 8 hypotheses. O-1 is capable of retrieving the oracle hypothesis with a single hypothesis used in its training, and attains best performance threshold with the inclusion of 1-best.

To further evaluate the performance difference between actual and oracle WER, the O-1 and EMBR  models were evaluated at each training checkpoint as shown in Figure~\ref{fig:numhyp}C. O-1's 1-best WER gets closer to the oracle WER as training progresses and matches the oracle performance at 100K iterations, indicating that use of all n-best hypotheses is unnecessary.

\begin{figure}[!tbh]
    \centering
    \includegraphics[trim={0cm 0cm 1cm 1cm},scale=0.4]{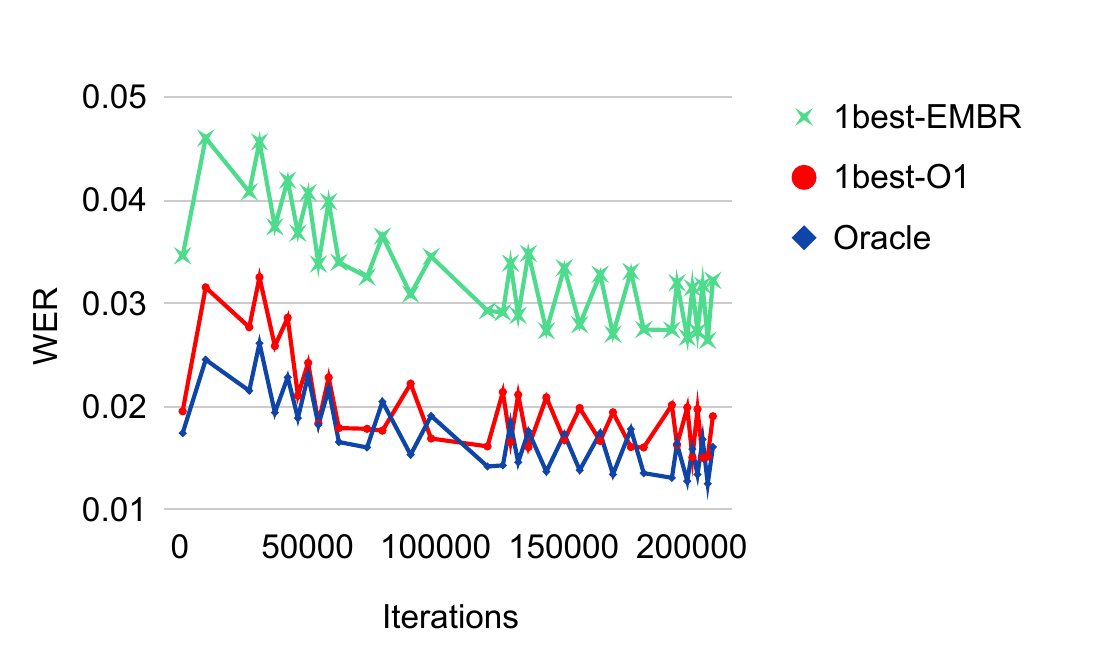}
    \caption{Comparison of 1best WER with oracle WER on librispeech test set using our proposed O-1 approach, baseline and EMBR}
    \label{fig:1bestmeetsoracle}
\end{figure}

\subsection{Results on O-1 Unsupervised Learning}
The experiments on a large-scale in-house data set comprising of short utterances up to 4 seconds long, shows a similar trend to the public data sets. Two audio sources were used for this evaluation, real and TTS generated audio. The test sets include several named entities. 
 For example, the RPN datasets contains proper nouns, place names, music genres, artists and other media related entities.
Each test set contains 10K utterances.
From Table~\ref{tab:names} we can see that O-1 attains 8\% and 9\% relative improvement over EMBR on two of the four testsets while yielding modest improvements on the remaining test sets. This indicates that the O-1 objective handles streaming recognition well compared to EMBR and scales to large-scale data sets.
\begin{table}[!tbh]
\centering
    \caption{Recognition performance on in-house data sets in U.S. English including VS and rare proper nouns (RPN-) from multiple application sources.}
    \label{tab:names}
\resizebox{7.5cm}{0.9cm}{
\begin{tabular}{ c c c c c c c }
\toprule 
\% WER & VS & RPN-M & RPN-N & \multicolumn{1}{c|}{RPN-P} & RPN-Q & RPN-Y\tabularnewline
\hline
\midrule
Baseline & 6.0 & 13.2 & 10.1 & 37.3 & 20.6 & 24.1\tabularnewline
EMBR & 5.8 & 13.1 & 9.9 & 36.9 & 20.3 & 24.0\tabularnewline
O-1 & 5.8 & 12.4 & 9.6 & 35.5 & 18.5 & 22.2\tabularnewline
\bottomrule 
\end{tabular}}
\end{table}
\section{Conclusion}
We have proposed O-1, a novel self-training objective to learn from oracle hypothesis. Our experiments shows that the O-1 reduces the difference between 1-best and oracle performance by a significant margin in multiple evaluation sets. O-1  shows promise by performing well on tasks where EMBR fails to improve. O-1 enjoys additional benefit of optimized training time over EMBR. In addition the results show that O-1 shows gains in both supervised and unsupervised setting under various test conditions. O-1 achieves 13\%, 15\% and 25\% relative improvement over EMBR on AMI-HM, Librispeech-testother and WSJ-eval92 testsets respectively. 
In a real-time, streaming setting, O-1 provides 9\% relative improvement over EMBR on a large scale in-house test set. O-1 can be further improved in the future using multiple hypotheses from the teacher model and applying advanced distillation approaches.    

\bibliographystyle{ieeebib_3auth}
\bibliography{ref_new}

\begin{thebibliography}{10}

\bibitem{wei2021theoretical}
Colin Wei, Kendrick Shen, et~al.,
\newblock ``Theoretical analysis of self-training with deep networks on
  unlabeled data,''
\newblock in {\em ICLR}, 2021.

\bibitem{grandvalet2004semi}
Yves Grandvalet and Yoshua Bengio,
\newblock ``Semi-supervised learning by entropy minimization,''
\newblock {\em Advances in NeurIPS}, vol. 17, 2004.

\bibitem{RanzatoCAZ15}
Marc'Aurelio Ranzato, Sumit Chopra, et~al.,
\newblock ``Sequence level training with recurrent neural networks,''
\newblock in {\em ICLR}, 2016.

\bibitem{williams1989learning}
Ronald~J Williams and David Zipser,
\newblock ``A learning algorithm for continually running fully recurrent neural
  networks,''
\newblock {\em Neural computation}, vol. 1, no. 2, pp. 270--280, 1989.

\bibitem{bengio2015scheduled}
Samy Bengio, Oriol Vinyals, et~al.,
\newblock ``Scheduled sampling for sequence prediction with recurrent neural
  networks,''
\newblock {\em Advances in NeurIPS}, vol. 28, 2015.

\bibitem{Bahdanau2016AnAA}
Dzmitry Bahdanau, Philemon Brakel, et~al.,
\newblock ``An actor-critic algorithm for sequence prediction,''
\newblock {\em ArXiv}, vol. abs/1607.07086, 2016.

\bibitem{pmlr-v97-collobert19a}
Ronan Collobert, Awni Hannun, et~al.,
\newblock ``A fully differentiable beam search decoder,''
\newblock in {\em Proceedings of the 36th ICML}, Kamalika Chaudhuri and Ruslan
  Salakhutdinov, Eds. 09--15 Jun 2019, vol.~97 of {\em PMLR}, pp. 1341--1350,
  PMLR.

\bibitem{prabhavalkar2018minimum}
Rohit Prabhavalkar, Tara~N Sainath, et~al.,
\newblock ``Minimum word error rate training for attention-based
  sequence-to-sequence models,''
\newblock in {\em IEEE ICASSP}. IEEE, 2018, pp. 4839--4843.

\bibitem{lu21b_interspeech}
Liang Lu, Zhong Meng, et~al.,
\newblock ``{On Minimum Word Error Rate Training of the Hybrid Autoregressive
  Transducer},''
\newblock in {\em Proc. Interspeech 2021}, 2021, pp. 3435--3439.

\bibitem{povey2005discriminative}
Daniel Povey,
\newblock {\em Discriminative training for large vocabulary speech
  recognition},
\newblock Ph.D. thesis, University of Cambridge, 2005.

\bibitem{muller-sennrich-2021-understanding}
Mathias M{\"u}ller and Rico Sennrich,
\newblock ``Understanding the properties of minimum {B}ayes risk decoding in
  neural machine translation,''
\newblock in {\em International Joint Conference on NLP (Volume 1: Long
  Papers)}, Online, Aug. 2021, pp. 259--272, Association for Computational
  Linguistics.

\bibitem{lu2021input}
Zhiyun Lu, Yanwei Pan, et~al.,
\newblock ``Input length matters: Improving rnn-t and mwer training for
  long-form telephony speech recognition,''
\newblock {\em arXiv preprint arXiv:2110.03841}, 2021.

\bibitem{goel2000minimum}
Vaibhava Goel and William~J Byrne,
\newblock ``Minimum bayes-risk automatic speech recognition,''
\newblock {\em Computer Speech \& Language}, vol. 14, no. 2, pp. 115--135,
  2000.

\bibitem{shannon2017optimizing}
Matt Shannon,
\newblock ``Optimizing expected word error rate via sampling for speech
  recognition,''
\newblock {\em arXiv preprint arXiv:1706.02776}, 2017.

\bibitem{mkbaskarprefix}
Murali~Karthick Baskar, Luk\'{a}\v{s} Burget, et~al.,
\newblock ``Promising accurate prefix boosting for sequence-to-sequence asr,''
\newblock in {\em IEEE ICASSP}, 2019, pp. 5646--5650.

\bibitem{weiran22_interspeech}
Wang Weiran, Tongzhou Chen, et~al.,
\newblock ``{Improving Rare Word Recognition with LM-aware MWER Training},''
\newblock in {\em Proc. Interspeech 2022}, 2022, pp. 1031--1035.

\bibitem{vesely2013sequence}
Karel Vesel{\`y}, Arnab Ghoshal, et~al.,
\newblock ``Sequence-discriminative training of deep neural networks.,''
\newblock in {\em Interspeech}, 2013, vol. 2013, pp. 2345--2349.

\bibitem{povey2001improved}
Daniel Povey and P~Woodland,
\newblock ``Improved discriminative training techniques for large vocabulary
  continuous speech recognition,''
\newblock in {\em ICASSP}. IEEE, 2001, vol.~1, pp. 45--48.

\bibitem{kingsbury2009lattice}
Brian Kingsbury,
\newblock ``Lattice-based optimization of sequence classification criteria for
  neural-network acoustic modeling,''
\newblock in {\em 2009 IEEE ICASSP}. IEEE, 2009, pp. 3761--3764.

\bibitem{povey2007evaluation}
Daniel Povey and Brian Kingsbury,
\newblock ``Evaluation of proposed modifications to mpe for large scale
  discriminative training,''
\newblock in {\em 2007 IEEE ICASSP'07}. IEEE, 2007, vol.~4, pp. IV--321.

\bibitem{fsdistill}
Mohammad Zeineldeen, Kartik Audhkhasi, et~al.,
\newblock ``Robust knowledge distillation from rnn-t models with noisy training
  labels using full-sum loss,''
\newblock in {\em IEEE ICASSP}, 2023, pp. 1--5.

\bibitem{rennie2017self}
Steven~J Rennie, Etienne Marcheret, et~al.,
\newblock ``Self-critical sequence training for image captioning,''
\newblock in {\em Proceedings of the IEEE conference on CVPR}, 2017, pp.
  7008--7024.

\bibitem{he2019exposure}
Tianxing He, Jingzhao Zhang, et~al.,
\newblock ``Exposure bias versus self-recovery: Are distortions really
  incremental for autoregressive text generation?,''
\newblock {\em arXiv preprint arXiv:1905.10617}, 2019.

\bibitem{arun2010unified}
Abhishek Arun, Barry Haddow, et~al.,
\newblock ``A unified approach to minimum risk training and decoding,''
\newblock in {\em Proceedings on Statistical Machine Translation and
  MetricsMATR}, 2010, pp. 365--374.

\bibitem{wiseman-rush-2016-sequence}
Sam Wiseman and Alexander~M. Rush,
\newblock ``Sequence-to-sequence learning as beam-search optimization,''
\newblock in {\em Proceedings of the 2016 Conference on EMNLP}, Austin, Texas,
  Nov. 2016, pp. 1296--1306, Association for Computational Linguistics.

\bibitem{cui21b_interspeech}
Xiaodong Cui, Brian Kingsbury, et~al.,
\newblock ``{Reducing Exposure Bias in Training Recurrent Neural Network
  Transducers},''
\newblock in {\em Proc. Interspeech 2021}, 2021, pp. 1802--1806.

\bibitem{bahdanau2016end}
Dzmitry Bahdanau, Jan Chorowski, et~al.,
\newblock ``End-to-end attention-based large vocabulary speech recognition,''
\newblock in {\em IEEE ICASSP}. IEEE, 2016, pp. 4945--4949.

\bibitem{Guo2020}
Jinxi Guo, Gautam Tiwari, et~al.,
\newblock ``Efficient minimum word error rate training of rnn-transducer for
  end-to-end speech recognition,''
\newblock in {\em Interspeech 2020}, 2020.

\bibitem{mbrdec2003}
V.~Doumpiotis, S.~Tsakalidis, et~al.,
\newblock ``Discriminative training for segmental minimum bayes risk
  decoding,''
\newblock in {\em ICASSP Proceedings}, 2003, vol.~1, pp. I--I.

\bibitem{sabour2018optimal}
Sara Sabour, William Chan, et~al.,
\newblock ``Optimal completion distillation for sequence learning,''
\newblock in {\em ICLR}, 2019.

\bibitem{graves2012sequence}
Alex Graves,
\newblock ``Sequence transduction with recurrent neural networks,''
\newblock {\em arXiv preprint arXiv:1211.3711}, 2012.

\bibitem{sutton1999policy}
Richard~S Sutton, David McAllester, et~al.,
\newblock ``Policy gradient methods for reinforcement learning with function
  approximation,''
\newblock {\em NeurIPS}, vol. 12, 1999.

\bibitem{baskar2022ask2mask}
Murali~Karthick Baskar, Andrew Rosenberg, et~al.,
\newblock ``Ask2mask: Guided data selection for masked speech modeling,''
\newblock {\em IEEE JSTSP}, vol. 16, no. 6, pp. 1357--1366, 2022.

\bibitem{park2020improved}
Daniel~S. Park, Yu~Zhang, et~al.,
\newblock ``{Improved Noisy Student Training for Automatic Speech
  Recognition},''
\newblock in {\em Proc. Interspeech 2020}, 2020, pp. 2817--2821.

\bibitem{hwang2022comparison}
Dongseong Hwang, Khe~Chai Sim, et~al.,
\newblock ``Comparison of soft and hard target rnn-t distillation for
  large-scale asr,''
\newblock {\em arXiv preprint arXiv:2210.05793}, 2022.

\bibitem{chan2021speechstew}
William Chan, Daniel Park, et~al.,
\newblock ``Speechstew: Simply mix all available speech recognition data to
  train one large neural network,''
\newblock {\em arXiv preprint arXiv:2104.02133}, 2021.

\bibitem{narayanan2019recognizing}
Arun Narayanan, Rohit Prabhavalkar, et~al.,
\newblock ``Recognizing long-form speech using streaming end-to-end models,''
\newblock in {\em 2019 IEEE ASRU}. IEEE, 2019, pp. 920--927.

\bibitem{gulati2020conformer}
Anmol Gulati, James Qin, et~al.,
\newblock ``Conformer: Convolution-augmented transformer for speech
  recognition,''
\newblock {\em arXiv preprint arXiv:2005.08100}, 2020.

\bibitem{park2019specaugment}
Daniel~S. Park, William Chan, et~al.,
\newblock ``{SpecAugment: A Simple Data Augmentation Method for Automatic
  Speech Recognition},''
\newblock in {\em Proc. Interspeech 2019}, 2019, pp. 2613--2617.

\bibitem{variani2020hybrid}
Ehsan Variani, David Rybach, et~al.,
\newblock ``Hybrid autoregressive transducer (hat),''
\newblock in {\em IEEE ICASSP}. IEEE, 2020, pp. 6139--6143.

\bibitem{kim2017generation}
Chanwoo Kim, Ananya Misra, et~al.,
\newblock ``Generation of large-scale simulated utterances in virtual rooms to
  train deep-neural networks for far-field speech recognition in google home,''
\newblock in {\em Interspeech}, 2017.

\bibitem{kudo2018sentencepiece}
Taku Kudo and John Richardson,
\newblock ``Sentencepiece: A simple and language independent subword tokenizer
  and detokenizer for neural text processing,''
\newblock {\em arXiv preprint arXiv:1808.06226}, 2018.

\bibitem{BaevskiZMA20}
Alexei Baevski, Yuhao Zhou, et~al.,
\newblock ``wav2vec 2.0: {A} framework for self-supervised learning of speech
  representations,''
\newblock in {\em Advances in NeurIPS}, 2020.

\bibitem{jouppi2021ten}
Norman~P Jouppi, Doe~Hyun Yoon, et~al.,
\newblock ``Ten lessons from three generations shaped google{'}s tpuv4i:
  Industrial product,''
\newblock in {\em 2021 ACM/IEEE ISCA}. IEEE, 2021, pp. 1--14.

\end{thebibliography}

\end{document}